\pdfoutput=1
\RequirePackage{fix-cm}
\documentclass[smallextended]{svjour3}       
\smartqed  
\usepackage{graphicx}
\usepackage{amssymb}
\usepackage{latexsym}
\usepackage{caption}
\usepackage{verbatim}
\usepackage{algorithm}
\usepackage{algorithmic}
\usepackage{tabularx}
\usepackage{amssymb}
\usepackage{amsmath}
\usepackage{natbib}

\begin{document}

\title{Generalizing the Convolution Operator in Convolutional Neural Networks
}


\author{Kamaledin Ghiasi-Shirazi
}


\institute{Kamaledin Ghiasi-Shirazi \at
              Department of Computer Engineering, Ferdowsi University of Mashhad (FUM), Office No.: BC-123, Azadi Sq., Mashhad, Khorasan Razavi, Iran. \\
              Tel.: +98-513-880-5158\\
              Fax: +98-513-880-7181\\
              \email{k.ghiasi@um.ac.ir}\\
              ORCID: 0000000160431820           
}

\date{Received: date / Accepted: date}

\maketitle

\begin{abstract}
Convolutional neural networks have become a main tool for solving many machine vision and machine learning problems. 
A major element of these networks is the convolution operator which essentially computes the inner product between a weight vector and the vectorized image patches extracted by sliding a window in the image planes of the previous layer.
In this paper, we propose two classes of surrogate functions for the inner product operation inherent in the convolution operator and so attain two generalizations of the convolution operator.
The first one is the class of positive definite kernel functions where their application is justified by the  kernel trick. 
The second one is the class of similarity measures defined based on a distance function. We justify this by tracing back to the basic idea behind the neocognitron which is the ancestor of CNNs.
Both methods are then further generalized by allowing a monotonically increasing function to be applied subsequently. 
Like any trainable parameter in a neural network, the template pattern and the parameters of the kernel/distance function are trained with the back-propagation algorithm. As an aside, we use the proposed framework to justify the use of sine activation function in CNNs. 
Our experiments on the MNIST dataset show that the performance of ordinary CNNs can be achieved by generalized CNNs based on weighted L1/L2 distances, proving the applicability of the proposed generalization of the convolutional neural networks. 
\end{abstract}

\section{Introduction}

The idea of using correlation/convolution operators in neural networks goes back to Fukushima who proposed cognitron \citep{fukushima1975cognitron} and neocognitron \citep{fukushima1980neocognitron,fukushima1988neocognitron} neural networks. In neocognitron, the input image is matched against a set of patterns that are represented by some weights. For each pattern, a map is produced in which the positions in the input image at which that very pattern is present are marked. 
This operation can be interpreted both as \textit{correlation} and \textit{convolution}, depending on how we assemble the weight vector as an image pattern. 
In this paper, we strive to the term \textit{convolution} since it is well established in this context and since correlational neural networks refer to a completely different network\citep{chandar2016correlational}.
To reduce the sensitivity to the exact positions of the patterns in the input image, Fukushima proposed sub-sampling of the produced maps with max-pooling.
By repeating these layers of convolution and pooling (which he termed U-layers and S-Layers) it was possible to represent more complex patterns by neurons at deeper layers. It is interesting that the convolution and max-pooling operations which constitute the backbone of today's convolutional neural networks (CNN) have been present in the very early design of neocognitron. Neocognitron was trained with an unsupervised learning algorithm that clearly implemented a matched filter. 

After the invention of the back-propagation algorithm by \citet{williams1986learning}, \citet{lecun1989backpropagation} introduced  CNNs which was essentially a neocognitron trained with the back-propagation algorithm. However, in contrast to the initial goal of implementing matched filters, it was observed that the weights of a trained CNN contained both positive and negative values. Having weights with negative values, the interpretation of convolution operator in CNNs generalized from the original idea of implementing a matched filter to a new interpretation in which the convolution operator extracts features from the image planes of the previous layer.
Whether we accept the matched filter or feature extraction viewpoints, the convolution operator is computing the inner product between image patches and a pattern represented by the weights. We call this the \textit{generalized matched filter viewpoint} in which we still view the inner product operator as a similarity measure, even though the pattern may take negative values. 
  
In this paper, we propose two generalizations of the convolution operator. 
In the first generalization, which is {based on} the kernel methods, we propose substituting the inner product operator within the convolution operation by a positive definite kernel function. 
In contrast to kernel methods such as support vector machines (SVM) and kernel principal component analysis (KPCA), here the positive definiteness of the kernel function is not crucial and we show that any monotonically increasing function of a positive definite kernel function can be used as well. 
The second generalization comes from the fact that the primary goal of including the convolution layer in the neocognitron and CNNs was to detect spots in the input plane that are locally similar to a target pattern. In this view, we propose that the inner product operation within the convolution operator can be replaced by a similarity measure. Specifically, we define a similarity measure as any monotonically increasing function of the negation of a distance function which assigns similarity zero to distance infinity. 
In this way, numerous similarity measures can be constructed by applying different monotonically increasing nonlinear functions to the negation of a distance metric. Therefore, instead of implementing a full similarity-based convolution layer, we implement a generalized convolution layer with the inner product operation replaced by the desired distance function. 
We then negate the result and apply an appropriate monotonically increasing function to arrive at a similarity measure. 
We have implemented our generalized convolution operators as layers within the well-known Caffe\citep{jia2014caffe} framework. 

Prior to this work, some researchers also proposed networks in which the inner product operation within the convolution operator had been replaced by some other function $f$. 
We denote the resulting operation as $f$-convolution.
\citet{serre2007robust} proposed the HMax model for object recognition where, in its second scale, the similarity between image patches and stored patterns are measured by a Gaussian function. In other words, the second scale of HMax computes a Gaussian-convolution. However, the HMax model is considerably different from a CNN and is not trained by the back-propagation algorithm.
In convolutional kernel networks (CKN), \citet{mairal2014convolutional} considered a special kernel function for measuring the similarity between two complete images and showed that its associated feature map can be approximated by a Gaussian-convolution followed by pooling. 
Assuming that the image patches and the learned patterns are normalized, \citet{mairal2014convolutional} showed that the computation of a Gaussian-convolution is equivalent to an ordinary convolution operator followed by a special nonlinearity that resembles the rectified linear unit (ReLU) in the interval [-1,1]. 
In case CKNs are implemented with the ordinary convolution operator followed by a nonlinearity, \citet{mairal2016} could train the network by the back-propagation algorithm. 
\citet{lin2014network} were the first who  explicitly proposed generalizing the convolution operator in CNNs and training the whole network with back-propagation.
They introduced the network in network (NIN) model in which the inner product operation within the convolution operator is replaced with a multilayer perceptron (MLP).
At first sight, considering the universal approximation property of multilayer perceptrons\citep{hornik1989multilayer} one may view NIN as a radical generalization of the convolution operator in which the internal inner product operation is substituted by an arbitrary function.
However, it can be shown that MLP-convolution is equivalent to an ordinary convolution followed by several $1\times 1$ convolutions and nonlinearities. 
In fact, this feature helped \citet{lin2014network} to implement NIN by ordinary CNNs without altering their implementation.
In this view, it can be said that the NIN did not generalize the convolution operator at all and all it did was the discovery that CNNs should be much deeper and should have a slow pace of decreasing the resolution of the convolution planes by pooling.
Although the broad idea of generalizing the convolution operator has been present in the above-mentioned works, the specific ideas presented in this paper are completely novel and it is for the first time that a network with a really generalized convolution operator is trained by the back-propagation algorithm.

The paper proceeds as follows. In section~\ref{sec:proposed-method}, we introduce two classes of generalized convolution operators. Some specific examples of generalized convolution operators are introduced in section~\ref{sec:example-of-generalized-convolution}. We found that simple random initialization of the parameters or applying algorithms like Xavier\citep{glorot2010understanding} are not suitable choices for initializing generalized convolutional neural networks (GCNN). Two initialization algorithms that can be used for the initialization of GCNNs are introduced in section~\ref{sec:initialization-methods}. We report our experiments on the MNIST dataset in section~\ref{sec:experiments}. We conclude the paper in section~\ref{sec:conclusions} and mention the future works in section~\ref{sec:future_works}.

\section{The proposed method}\label{sec:proposed-method}
CNN is a deep neural network which consists of different types of layers, including convolution, pooling, nonlinearity, inner product, and loss layers. Usually, a module consisting of a sequence of convolution, nonlinearity, and pooling layers is repeated several times to produce a suitable representation of the input data which is then fed to a fully connected network to estimate the output (for recent generalizations of this block see \citep{szegedy2015going,he2016deep}). 
The convolution layer computes the convolution between its input planes and several filters represented by the weight parameters and produces a set of output planes, one associated with each filter. However, the convolution is essentially a linear operation.
It is well-known that deepening of neural networks with linear activation function does not increase the representational power and these networks are still representing a linear function of the input data.
So, to increase the modeling capability of CNNs, the convolution layer is usually followed by a nonlinearity layer. Some examples of common activation functions include rectified linear unit (ReLU), tangent hyperbolic(TanH), and logistic sigmoid (Sigmoid).
The goal of the pooling layer is to reduce the sensitivity of the network to translations of the input images and to summarize the important information of the input planes in a more compact representation.

Consider a convolution layer which operates on $p$ input planes and generates $q$ output layers. Assume that $I^{(i,j)}$ is an n-dimensional vector generated by vectorizing $p$ patches of input planes centered at position $(i,j)$. Let $W^\ell$ be the weight vector associated with $\ell$'th output plane.
Then the output value at position $(i,j)$ of plane $\ell$ is computed by formula

\begin{equation}
O^\ell_{i,j} = \sum_{k=1}^n {I^{i,j}_k W^\ell_k} = \langle I^{i,j}, W^\ell \rangle \label{eq:inner_product_conv}
\end{equation}

Our first proposal for generalizing the convolution operator is to replace the inner product operation in Eq.(\ref{eq:inner_product_conv}) with a positive definite kernel function. Assuming that $k$ is a positive definite kernel function, the output is now computed by

\begin{equation}
O^\ell_{i,j}  = k( I^{i,j}, W^\ell) \label{eq:kernel_conv}
\end{equation}

This generalization allows us to use a handful of kernel functions such as Gaussian, polynomial, Laplacian, cosine, Cauchy, and intersection in place of the inner product operation.
However, our choice of using kernel functions is severely restricted by the positive definiteness requirement. In a kernel method like SVM, the positive definiteness property plays a crucial role and violation of it makes the objective function unbounded from below. In RBF neural networks, the positive definiteness property of kernel functions guarantees that the kernel matrix would be invertible and so the optimal weights of radial basis functions exist and are unique. 
Generally, the positive definiteness property of kernel functions eliminates the possibility that the inner product of a vector with itself {becomes} negative. 
Assume that $k$ is a inner product kernel function with feature space $F$ and feature map $\phi$ (i.e. $k(x,z)=\langle\phi(x),\phi(z)\rangle_F$) which is not positive definite.
Since $k$ is not positive definite, there exist inputs $x_1,...,x_n$ and coefficients $\alpha_1,...,\alpha_n$ such that

\begin{equation}
\sum_{i=1}^n\sum_{j=1}^n \alpha_i\alpha_j k(x_i,x_j) < 0 \label{eq:kernel_non_pos_def}
\end{equation}

One can easily verify that the expression on the left side of Eq.(\ref{eq:kernel_non_pos_def}) is equal to inner product of the vector $\alpha_1 \phi(x_1) + ... + \alpha_n \phi(x_n)$ with itself. 
Therefore, use of non-positive definite kernel functions in GCNNs may lead to patterns which are not similar to themselves, violating the \textit{generalized pattern matching viewpoint}. However, in GCNNs we are not concerned with all vectors that can be constructed in the feature space associated with a kernel function. Instead, we are applying the kernel function directly to two input vectors and the requirement of similarity of patterns to themselves translates to the condition $k(x,x)\geq 0$ for all input vectors x. This requirement is satisfied for any function $k$ having the form $k(x,z)=f(k'(x,z))$, where $k'$ is a positive definite kernel function and $f$ is a monotonically increasing function with $f(0)\geq 0$. So, we {arrive at} our first generalization of the convolution operator.

\textbf{Generalization 1:} The convolution operator in CNNs can be generalized by substituting the inner product operation $x^T w$ between a vectorized input $x$ and a weight vector $w$ by $f(k(x,w))$, where $k$ is a positive definite kernel function and $f$ is a monotonically increasing function with $f(0)\geq 0$.

It is evident that the main purpose of the inner product stage within the convolution operator in neocognitron was to measure the similarity between the patches of the input maps of the preceding layer and a template pattern. One justification for this is that the inner product operator is essentially a similarity measure \citep[see section 1.1 of][]{Scholkopf_2002}. It may be argued that the inner product operation is not a suitable similarity measure since for example there are vectors which are more similar to a chosen vector than itself.
Our second proposal for generalizing the convolution operator is to substitute the inner product operator with a similarity measure. Defining similarity based on a distance measure, we {arrive at} our second generalization of the convolution operator. 

\textbf{Generalization 2:} The convolution operator in CNNs can be generalized by substituting the inner product operation $x^T w$ between a vectorzied input $x$ and a weight vector $w$ by $f(-d(x,w))$, where $d$ is a distance metric and $f:(-\infty,0]\to [0,\infty)$ is a monotonically increasing function with $f(-\infty)=0$.  Adding the additional constraint $f(0)=1$ ensures that $f(-d(x,x))=1$, meaning that the similarity of each vector $x$ with itself is 1. 
However, since the dimension of the input space is usually high\footnote{For example, in our experiments on the MNIST dataset, we have 12 planes in the first convolution layer which, considering a window of size 5, induces a dimensionality of $12\times 5\times 5=300$ on the input of the second convolution layer.}, exact matching almost never happens and even similar items typically have high numerical distances. In addition, since we want to permit the use of non-squashing activation functions such as ReLU, which have experimentally proven to do better than squashing functions such as TanH and Sigmoid, we don't impose the restriction $f(0)=1$. 
 
In practice, we implemented similarity/kernel functions by using multiple layers in Caffe. These layers include a generalized convolution layer based on a metric distance(e.g. weighted L2 distance), possibly followed by an AdaptiveLinear layer with negative slope\footnote{AdaptiveLinear is a simple new kind of layer that we have added to Caffe which implements $y=ax+b$, where the parameters $a$ and $b$ differ between output channels.}, followed by an activation function layer like exponential (Exp) or ReLU. In this view, activation functions are appropriate monotonically increasing functions that complement the functionality of a distance-based generalized convolution layer such that the whole module implements a generalized convolution operator. 
For example, ReLU activation function can be seen as a monotonically increasing function that complements the role of the preceding layers by enforcing the non-negativity criterion of a similarity measure. 

\section{Examples of generalized convolution operators}\label{sec:example-of-generalized-convolution}

\subsection{Non-isotropic Gaussian kernel}\label{sec:theory-gaussian}
Non-isotropic Gaussian kernel is defined as 

\begin{equation}
k(x,z)=\exp\left(-\frac{1}{2}\sum_{i=1}^n \tau_i \left(x_i-z_i\right)^2 \right) \label{eq:kernel_gaussian}
\end{equation} 
where $\tau$ is the precision vector which consists of positive values. Use of Gaussian kernel function in generalized convolution is admissible since both it is a positive definite kernel function and it can be expressed as the application of the monotonically increasing function $f:(-\infty,0]\to[0,\infty)$ with the definition $f(x)=\exp(-\frac{1}{2}x^2)$ to the negation of the weighted L2 distance(WL2Dist).
In addition to satisfying the required constraint $f(-\infty)=0$, the $\exp(-\frac{1}{2}x^2)$ function has the additional property that $f(0)=1$, ensuring that it is a similarity measure spanning the range $(0,1]$. 

\subsection{Non-isotropic Laplacian kernel}
Non-isotropic Laplacian kernel is defined as 

\begin{equation}
k(x,z)=\exp\left(-\sum_{i=1}^n \tau_i \left|x_i-z_i\right| \right) \label{eq:kernel_laplacian}
\end{equation}
where $\tau$ is the precision vector which should be positive. Again, Laplacian kernel function is both a positive definite kernel function and it can be expressed as the application of the monotonically increasing function $\exp(x)$ to the negation of the weighted L1 distance(WL1Dist). 

\subsection{Cosine kernel: justifying the Sine activation function}\label{sec:theory-sine}
It is well known that $k(x,z)=cos(w^T(x-z))$ is a positive definite kernel function and so it can be used to measure the similarity between an input image patch $x$ and a pattern $z$. Since $w$ is the parameter of the kernel, it is fixed and $-w^T z$ is equal to some constant $b'$. It follows that $k(x,z)=cos(w^T x + b')=sin(w^T x + b)$, where $b=b'+\pi/2$. 
Thus, $sin(w^T x + b)$ is essentially computing the kernel function $k(x,z)=cos(w^T(x-z))$, where z is an implicit pattern satisfying the equation $b=-w^T z+\pi/2$. Note that in contrast to the ordinary convolution operation where the similarity of $x$ is measured against the vector of weights $w$, here $w$ is solely a parameter of the kernel function and the desired pattern $z$ is hidden in the bias parameter $b$. 
The above line of reasoning works exactly for the cosine activation function. However, since the gradient of the cosine function vanishes at zero, the parameters of a network with cosine activation function would get stuck in their initial values. This is because almost all initialization algorithms initialize the weight vector $w$ and the bias parameter $b$ in a way that $w^T x + b$ is on average zero.

\begin{figure}[h]
\hfill
\begin{center}
\includegraphics[width=3.4in]{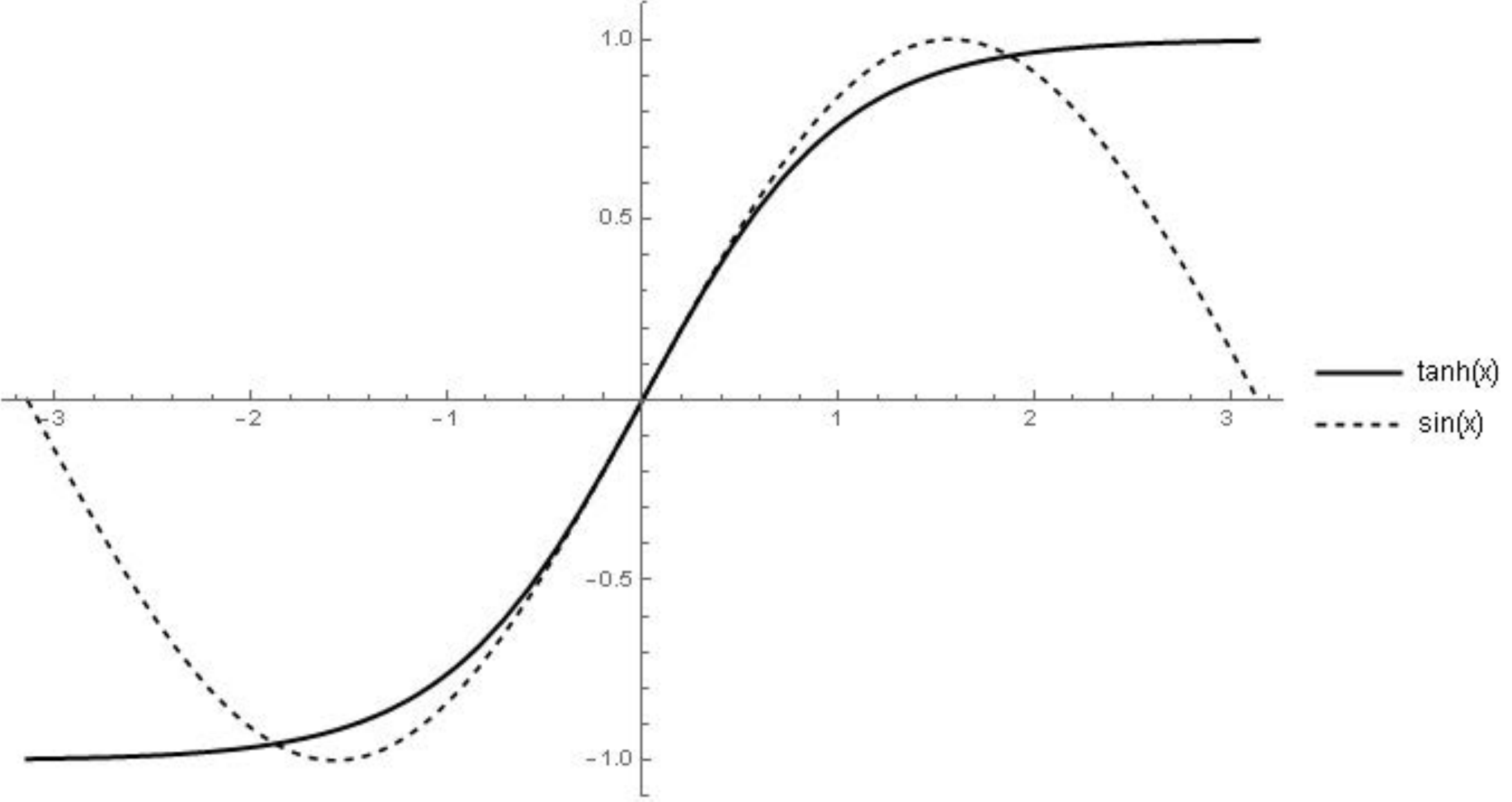}
\end{center}
\caption{Sine and TanH activation functions. The graphs of these functions are close to each other in $[-\pi/2,\pi/2]$}.
\label{Fig:sine_vs_tanh}
\end{figure}

In section~\ref{sec:experiment-sine} we will experimentally show that the Sine activation function works similar to ReLU and significantly better than TanH. 
One benefit of Sine is that it does not have the saturation problem of TanH and Sigmoid.
As is illustrated in Figure~\ref{Fig:sine_vs_tanh}, Sine and TanH have similar shapes in the range $[-\pi/2,\pi/2]$, however, outside this region TanH is saturated while Sine is periodic. 
One problem with TanH is that if the target value is $1$, then the weights are pushed towards infinity and the gradient of the TanH function vanishes.
To remedy this problem, \citet{lecun1998efficient} proposed a scaled version of TanH with definition

\begin{equation}
y=1.7159 \tanh \left(\frac{2}{3} x\right) \label{eq:scaled_tanh}
\end{equation}.

Another benefit of Sine is that it can produce an output value of $1$ without pushing the weights towards infinity. 

\subsection{Other similarities based on the weighted L2-distance}\label{sec:WL2Dist}
We saw in section~\ref{sec:theory-gaussian} that Gaussian kernel equals to the composition of the monotonically increasing function $f:(-\infty,0]\to[0,\infty)$ with definition $f(x)=exp(-\frac{1}{2}x^2)$ and the negation of WL2Dist. In this section, we propose to use other activation functions on top of the WL2Dist.
For the case of Gaussian kernel, we can view Eq.~(\ref{eq:kernel_gaussian}) as the unnormalized Gaussian probability density that an input patch $x$ matches pattern $z$. Rewriting Eq.~(\ref{eq:kernel_gaussian}) in the usual form of a multivariate Gaussian distribution we have

\begin{equation}
k(x,z)\propto \frac{|J|^{\frac{1}{2}}}{(2\pi)^{\frac{n}{2}}} \exp\left(-\frac{1}{2} (x-z)^T J (x-z)\right) \label{eq:kernel_gaussian_information}
\end{equation}
where $J$ is a diagonal precision matrix with entries $\tau_1,...,\tau_n$.

Viewing the Gaussian kernel as a probability density function, we can say that the value returned by this function is proportional to the probability density of input $x$ in a Gaussian distribution with mean $z$ and precision matrix $J$. 
The problem with this value is that it cannot directly be used as a measure for deciding whether the input data is similar to the desired pattern $z$ or not. 
In this section, we exploit this probabilistic point of view to {arrive at} some other activation functions on the top of the WL2Dist. 

Since, by assumption, the precision matrix of the Gaussian distribution is diagonal, it follows that different dimensions are independent of each other and so the squared weighted L2 distance $D=\sum_{i=1}^n \tau_i \left(x_i-z_i\right)^2$ is equal to the sum of squares of $n$ standard normal variables which is known to have a $\chi^2$ distribution with $n$ degrees of freedom.
So, the probability that an input $x$ with $D\ge d$ belongs to the Gaussian distribution associated with pattern $z$ is equal to the p-value of the $\chi^2$ distribution at point $d$. Using {$\chi^2$ inverse cumulative distribution}, one can determine two thresholds $D_L$ and $D_H$ such that for inputs $x$ with $D(x,z)<D_L$ the probability that $x$ is generated by this distribution is very high and for inputs $x$ with $D(x,z)>D_H$ this probability is very low. Therefore, a suitable value for similarity of input $x$ to pattern $z$ is given by

\begin{eqnarray*} 
sim(x,z) =   \left\{ \begin{array}{c}
			1\;\;\;\;\;\;\;\;\; if\;D(x,z)\le D_L \\
			\frac{D_H-D(x,z)}{D_H-D_L}\;\;\;\;\;\;\; if\;D_L<D(x,z)\le D_H \\
			0\;\;\;\;\;\;\;\;\; if\;D(x,z)>D_H \end{array} \right.
\end{eqnarray*}

If we first apply a linear transformation with the slope $\frac{-1}{D_H-D_L}$ and the bias $\frac{D_H}{D_H-D_L}$ to the output of the WL2Dist layer, then the desired functionality can be achieved using the DoubleThreshold activation function defined as

\begin{eqnarray*} 
f(x) =   \left\{ \begin{array}{l}
			0\;\;\;\; if\;x< 0 \\
			x\;\;\;\; if\;0\le x< 1 \\
			1\;\;\;\; if\;x\ge 1 \end{array} \right.
\end{eqnarray*}



If we allow similarities greater than $1$, then the upper limit of the DoubleThreshold activation function is dropped and we {reach at} the well-known ReLU activation function.

\subsection{Other similarities based on the weighted L1-distance}\label{sec:WL1Dist}
The discussion of the previous section can be repeated for the Laplacian distribution, resulting in generalized convolution operators based on the WL1Dist. 

\section{Initialization of parameters}\label{sec:initialization-methods}
\begin{figure}[h]
\hfill
\begin{center}
\includegraphics[width=3.4in]{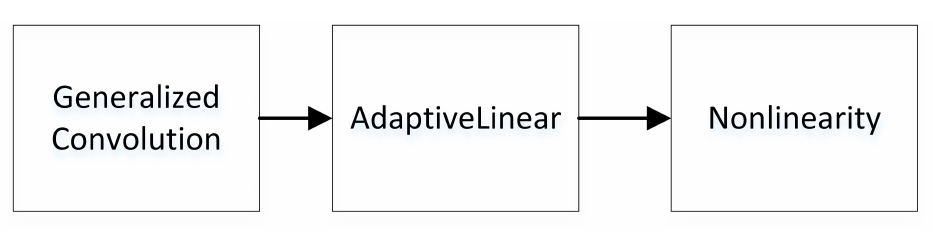}
\end{center}
\caption{General form of the sequence of layers that implement a generalized convolution operator. }
\label{Fig:InitializationModel}
\end{figure}

The performance of deep CNNs is strongly influenced by the method of initializing the parameters and by controlling the amount of backward gradient returned to each parameter\citep{krahenbuhl2015data, mishkin2015all, glorot2010understanding, ioffe2015batch}. Initialization and optimization algorithms proposed for neural networks are designed based on the linear model of neurons (i.e. $y=w^T x+b$). When the incoming weights are initialized randomly with mean zero, this model ensures that the mean of the output is zero as well. By choosing appropriate values for the magnitudes of weights one can ensure that all neurons have a mean value of zero and a variance of one, avoiding the vanishing/exploding problems in the forward pass \citep{lecun1998efficient}. Recently, similar approaches have been devised that control the magnitude of the gradient in the backward pass \citep{glorot2010understanding, ioffe2015batch,krahenbuhl2015data}. However, by substituting the inner product operation with a kernel/distance function all of these nice properties fade and the vanishing/exploding problems reappear both in the forward and backward passes. 
Each kernel/distance function has its own properties that should be considered in initializing its parameters. In this section, we consider two initialization algorithms for networks based on the weighted L1/L2 distances. The specific architecture we are considering is depicted in Figure~\ref{Fig:InitializationModel}. 

\subsection{Precision adjustment initialization algorithm}
The goal of this initialization method is to ensure that all signals at the forward pass have appropriate magnitudes. The initialization proceeds module by module, adjusting the precision parameters of the generalized convolution layer of each module to ensure that the empirical mean of the signal {passed} to the subsequent nonlinearity layer is zero and its empirical variance is some target value $\sigma$. 
This algorithm is very similar to the initialization algorithm of \citet{mishkin2015all} except that here the precision parameters are pre-initialized randomly while \citet{mishkin2015all} pre-initialize weights with orthonormal matrices.
Since this algorithm only adjusts the precision parameters of the generalized convolution layers, we call it the \textit{precision adjustment} initialization algorithm. 
\subsection{Whole-network adjustment initialization algorithm}
\citet{glorot2010understanding} proposed that the weights should be initialized in a way that both the activation values of neurons in the forward pass and the gradient back-propagated in the backward pass have appropriate variances. They proposed an analytical algorithm for initializing the weights of a convolutional neural network based on this idea. Recently, \citet{krahenbuhl2015data} proposed a data-dependent iterative algorithm for attaining this goal and showed that their approach works superior to the analytical approach of \citet{glorot2010understanding}, at least in the experiments reported in their paper. {We added the support for AdaptiveLinear, WL1Dist, and WL2Dist layers to the implementation of \citet{krahenbuhl2015data}}. Since this algorithm adjusts the parameters of the whole network, layer by layer, we call it the \textit{whole-network adjustment} algorithm.

\section{Experiments}\label{sec:experiments}
In this section, we experimentally evaluate two realizations of GCNNs. 
To compare GCNNs with [ordinary] CNNs in the fairest and most informative way, we conduct our experiments on the MNIST\citep{lecun1998gradient} dataset which has been a classical testbed for CNNs from its advent till now. 
The MNIST dataset is a collection of 60000 training and 10000 testing samples of handwritten digits.
We train the networks with the official training samples, without applying any distortions. 
 
\subsection{Experimental setup}\label{sec:exp-setup}
\begin{figure}[h]
\hfill
\begin{center}
\includegraphics[width=3.4in]{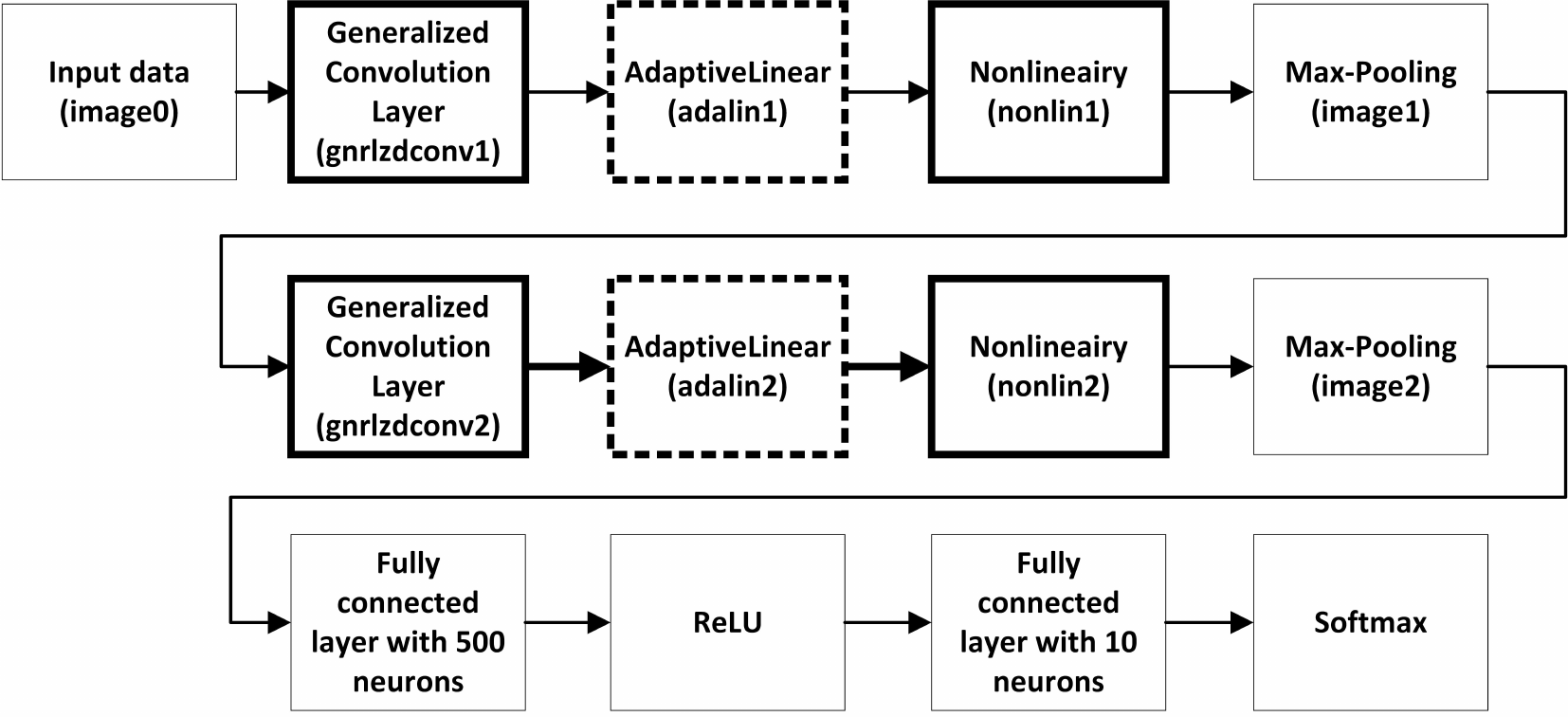}
\end{center}
\caption{General form of the network architecture of GCNNs chosen for our experiments on the MNIST dataset.}
\label{Fig:NetworkArchitecture}
\end{figure}

The general form of the network architecture considered in this section is depicted in Figure~\ref{Fig:NetworkArchitecture}. Only boxes with thick border may differ between the  experiments. The dotted boxes of the AdaptiveLinear layers imply that they may not be present in some experiments (or are present with slope 1, bias 0, and zero learning rate multipliers). The number of planes of the first and second generalized convolution layers is $12$ and $48$, respectively, with a window size of $5$. 
In our experiments, we set the base learning rate to $0.01$, the momentum to $0.9$, the mini-batch size to $100$, the maximum number of iterations to $18000$ (which is equivalent to $30$ epochs), and the weight decay coefficient to $0$. The learning rate at $n$'th iteration is computed by dividing the base learning rate by $(1+\gamma n)^p$, where $\gamma =0.0001$ and $p = 0.75$. 
We chose a significance level of $0.05$ for determining the statistical significance of the experiments. If not explicitly mentioned, the number of repetitions of each experiment is $25$. To ensure exact reproducibility  of the results, we have set the \textit{random\_seed} parameter of each experiment to a deterministic function of the experiment number. 

\subsection{Some insights into ordinary convolutional neural networks}
In this section, we perform some experiments on ordinary CNNs that will prove to be useful in design and analysis of our experiments on GCNNs. 
First, we conduct experiments to investigate the role of training of the convolution layers in the accuracy of CNNs. Second, we investigate the role of the negative weights in the accuracy of CNNs. 

\subsubsection{Investigating the role of convolution layers}
In this experiment, we want to identify the role of the convolution layers in the accuracy of CNNs on the MNIST dataset. In the experimental settings described in Section~\ref{sec:exp-setup}, an ordinary CNN obtains an accuracy of $99.138\%$ on the MNIST dataset. If we confine the weights of the convolution layers to their initial random values (by setting their associated learning rate multipliers to zero), the accuracy declines to $98.245\%$. This shows that only less than $0.9\%$ of the accuracy of a CNN on the MNIST dataset is due to training of the convolution layers.

\subsubsection{Studying matched filter CNNs}
In this experiment, our goal is to study the effect of negative weights on the accuracy of CNNs. In the matched filter viewpoint of CNNs, the weights of the convolution layer should represent a cluster of patches of the preceding layer. Since the values of the original image are between $[0,1]$ and the values of all convolution layers are passed from ReLU, all input maps to the convolution layers are positive. 
In this experiment, we want to examine the effect of confining the weights of the convolution layers to positive values on the accuracy of CNNs. 
To minimize the unwanted effects of initialization and optimization issues on this experiment, we force the positivity of the weights by the absolute value operation, so that the magnitudes of the gradients with respect to weights is unchanged. 
For a network with positive weights, the average activation of neurons would no longer be zero and the initialization algorithm of Xavier\citep{glorot2010understanding} is inapplicable. So, we first use the \textit{precision adjustment} initialization algorithm to ensure that the outputs of the convolution layers have mean zero and standard deviation $0.5$. 
After training for one epoch (i.e. 600 iterations), we use the whole-network adjustment initialization algorithm to initialize the weights, ensuring that the gradients returned to all layers have appropriate magnitude

To eliminate the effect of initialization algorithm, we first trained an ordinary CNN with the \textit{whole-network adjustment} and obtained an accuracy of $99.104\%$. 
After imposing the positivity constraint on weights, the accuracy significantly reduced to $98.817\%$. 
This proves that the negative weights have a remarkable role in the high accuracy of CNNs and the matched filter viewpoint (at least when weights are initialized randomly) cannot fully explain the high accuracy of CNNs.
We conclude from this experiment that in GCNNs we should also allow the template pattern to take negative values. For example in the case of distance-based generalized convolution operators (such as Gaussian, Laplacian, WL1Dist, and WL2Dist) we allow the mean parameter to take negative values. 

\subsubsection{Effect of initialization method on accuracy of ordinary CNNs}

None of the GCNNs considered in this paper can be trained with  randomly initialized weights. So, we ought to resort to the initialization algorithms of section~\ref{sec:initialization-methods}. In this section, we want to study the suitability of these initialization algorithms for initializing an ordinary CNN on the MNIST dataset. In addition to initialization algorithms of section~\ref{sec:initialization-methods}, we also consider the Xavier algorithm\citep{glorot2010understanding} which is the initialization algorithm chosen by the MNIST example in Caffe\footnote{Truly speaking, although \cite{glorot2010understanding} introduced a new algorithm which considers the backward gradient, the Xavier initialization algorithm in Caffe with default parameters is what has been introduced by \citet{lecun1998efficient} many years ago.}.  
Table~\ref{Table:EffectOfInitializationOnCNN} shows the accuracies obtained by different initialization algorithms on ordinary CNNs . 
As it can be seen, our data-dependent \textit{precision adjustment} algorithm works significantly better than both Xavier\citep{glorot2010understanding} and \textit{whole-network adjustment}\citep{krahenbuhl2015data} in this experiment. So, in section~\ref{sec:experiment-L1L2}, when comparing CNNs with GCNNs, we also consider CNNs initialized by the \textit{precision adjustment} algorithm.

\begin{table}[h]
\caption{Accuracies of CNNs initialized by different algorithms on the MNIST dataset.}
\begin{center}
\begin{tabular}{|l|r|}\hline
	Initialization algorithm & accuracy \\
\hline
   Xavier\citep{glorot2010understanding}  &   $99.138\pm0.058\%$ \\
   Precision adjustment  &   $\mathbf{99.180\pm0.070\%}$ \\
   
   Whole-network adjustment&   $99.104\pm0.065\%$ \\
   
   \hline
\end{tabular}
\end{center}
\label{Table:EffectOfInitializationOnCNN}
\end{table}

\subsection{Experimental evaluation of the Sine activation function}\label{sec:experiment-sine}
In section~\ref{sec:theory-sine} we showed that the use of Sine as a neural activation function can be explained from a kernel methods perspective. 
In this section, we experimentally compare Sine with other important activation functions such as Sigmoid, TanH , and ReLU. To identify the role of negative values at the output, we also include the rectified sine (ReSine) and rectified tangent hyperbolic (ReTanH) activation functions in our experiments. Table~\ref{Table:ComparisonOfActivationFunctions} shows the results of these experiments. In all experiments, we have used the Xavier\citep{glorot2010understanding} algorithm for initialization of the parameters. As the results show, Sine has performed significantly better than Sigmoid and TanH. However, the slightly higher accuracy of Sine in comparison to ReLU is not [statistically] significant.

\begin{table}[h]
\caption{Accuracies obtained by different activation functions applied to the output of convolution layers of a $12\times 48$ CNN on the MNIST dataset. All networks are initialized by the Xavier\citep{glorot2010understanding} algorithm.}
\begin{center}
\begin{tabular}{|l|r|}\hline
   Activation & Accuracy \\\hline
   None  &   $99.066\pm0.056\%$ \\
   Sigmoid  &   $98.774\pm0.057\%$ \\ 
   
   TanH &   $99.096\pm0.056\%$ \\   
   ReLU &   $99.138\pm0.058\%$ \\      
   ReTanH  &   $99.106\pm0.063\%$ \\
   ReSine  &   $99.120\pm0.072\%$ \\      
   Sine  &   $99.146\pm0.070\%$ \\
   \hline
\end{tabular}
\end{center}
\label{Table:ComparisonOfActivationFunctions}
\end{table}


\subsection{Weighted L1 and L2 distances}\label{sec:experiment-L1L2}

In sections~\ref{sec:WL2Dist} and \ref{sec:WL1Dist}, we showed that WL1Dist/WL2Dist+AdaLin+ReLU modules implement legitimate generalized convolution operators. In this section, we want to study these modules experimentally. We force the positivity of the precision parameters with absolute value operation. The mean parameter is initialized with a uniform distribution on the non-negative range $[0,1]$ but is allowed to take arbitrary positive/negative values during learning. 
We experimentally found that the maximum of $18000$ iterations chosen for ordinary CNNs is not sufficient for full training of GCNNs in this experiment and so increased this number to $36000$ iterations. We also repeated our previous experiments with ordinary CNNs with 36000 iterations which slightly improved the previous results. Each experiment is repeated with values $1$ and $10$ for learning rate multipliers (lr-mult) of the parameters of the generalized convolution layers. Table~\ref{Table:experiments-WL1-WL2} shows the accuracies obtained with different initialization algorithms. In each row, accuracies that are significantly higher than others are boldfaced.



\begin{table}[h]
\caption{Accuracies of ordinary CNNs, WL1Dist GCNNs, and WL2Dist GCNNs on the MNIST dataset averaged over $25$ runs. Each experiment is performed twice with values $1$ and $10$ for the learning rate multipliers (lr-mult) associated with the parameters of the generalized convolution layers. In this table PrecAdj stands for the precision adjustment algorithm and WhlNetAdj stands for the whole network adjustment algorithm.}
\begin{center}
\begin{tabular}{|l|c|c|c|c|}\hline
   Init.  alg.&   lr-mult & CNN & WL1Dist-GCNN & WL2Dist-GCNN  \\\hline
   Xavier  &   1  & $99.171\pm0.056\%$ & - & -\\
   
   Xavier  &  10  & $99.236\pm0.060\%$ & - & -\\

   PrecAdj  & 1&   $99.143\pm0.058\%$   & $99.156\pm0.064\%$ & $99.139\pm0.069\%$ \\

   PrecAdj  & 10&   $99.234\pm0.065\%$  & $99.218\pm0.111\%$ & $99.244\pm0.060\%$ \\
   WhlNetAdj  & 1&   $99.128\pm0.051\%$  & $\mathbf{99.173\pm0.055\%}$ & $99.138\pm0.061\%$\\

   WhlNetAdj  & 10&   $99.094\pm0.149\%$ & $99.084\pm0.110\%$ & $\mathbf{99.154\pm0.058\%}$\\
   \hline
\end{tabular}
\end{center}
\label{Table:experiments-WL1-WL2}
\end{table}

\section{Conclusions}\label{sec:conclusions}
In this paper, we proposed two methods for generalizing the convolution operator in CNNs. The first method is based on substituting the inner product operation within the convolution operator with a monotonically increasing function of a positive definite kernel function. In the second method, we replace the inner product operator with a monotonically increasing function of the negation of a distance function. 
In this paper, we implemented generalized CNNs (GCNN) based on the cosine kernel and weighted L1/L2 distances, and showed that the resulting networks achieve or even slightly surpass the accuracies of ordinary CNNs on the MNIST dataset.  
However, we believe that the main merit of this research is that it introduces a generalized conceptual framework that paves the way for the application of sophisticated methods developed in other fields of machine learning at the heart of GCNNs. Some of the machine learning methods that can be potentially used in GCNNs include kernel principal component analysis, multiple kernel learning, infinite kernel learning, metric learning, and similarity learning.
In addition, this work sheds more light on the nature of the convolution operator as a central element of CNNs.


\section{Future works}\label{sec:future_works}
In this paper, we introduced the key idea that the convolution operator in CNNs can be generalized by a wide class of kernel/distance functions. We experimentally supported this idea by implementing two generalized convolution operators based on the weighted L1/L2 distance functions and carrying out experiments on the MNIST dataset. In the future, we aim to study and improve the proposed approach in several directions. 
First, we plan to implement the weighted L1/L2 distance generalized GCNNs on GPU and apply them to more challenging datasets like CIFAR10 and CIFAR100\citep{krizhevsky2009learning}, and Imagenet\citep{ILSVRC15}. 
This would be a difficult task since the successful network models proposed for these datasets are very deep and use other complementary techniques, such as dropout\citep{srivastava2014dropout}, which are not yet adapted to the proposed generalized framework.
Second, we decide to exploit the discovered link between the kernel methods and CNNs to apply kernel methods machinary (such as SVM, KPCA, KFDA, MKL, and IKL) to CNNs. 
Finally, this work can be followed by implementing other possible forms of GCNNs(e.g. those based on polynomial or inverse multiquadric kernels). Our preliminary experiments suggest that almost every generalization of the convolution operator requires its own handling of initialization and optimization algorithms. 

\subsection*{Acknowledgments}
The author wishes to express appreciation to Research Deputy of Ferdowsi University of Mashhad for supporting this project by grant No.: 2/43037.
The author also thanks his fellows Ahad Harati and Ehsan Fazl-Ersi for their valuable comments.

\subsection*{References}
\bibliography{GeneralizedConvlolutionalNeuralNetworks}

\begin{thebibliography}{24}
\providecommand{\natexlab}[1]{#1}
\providecommand{\url}[1]{{#1}}
\providecommand{\urlprefix}{URL }
\expandafter\ifx\csname urlstyle\endcsname\relax
  \providecommand{\doi}[1]{DOI~\discretionary{}{}{}#1}\else
  \providecommand{\doi}{DOI~\discretionary{}{}{}\begingroup
  \urlstyle{rm}\Url}\fi
\providecommand{\eprint}[2][]{\url{#2}}

\bibitem[{Chandar et~al(2016)Chandar, Khapra, Larochelle, and
  Ravindran}]{chandar2016correlational}
Chandar S, Khapra MM, Larochelle H, Ravindran B (2016) Correlational neural
  networks. Neural computation

\bibitem[{Fukushima(1975)}]{fukushima1975cognitron}
Fukushima K (1975) Cognitron: A self-organizing multilayered neural network.
  Biological cybernetics 20(3-4):121--136

\bibitem[{Fukushima(1980)}]{fukushima1980neocognitron}
Fukushima K (1980) Neocognitron: A self-organizing neural network model for a
  mechanism of pattern recognition unaffected by shift in position. Biological
  cybernetics 36(4):193--202

\bibitem[{Fukushima(1988)}]{fukushima1988neocognitron}
Fukushima K (1988) Neocognitron: A hierarchical neural network capable of
  visual pattern recognition. Neural networks 1(2):119--130

\bibitem[{Glorot and Bengio(2010)}]{glorot2010understanding}
Glorot X, Bengio Y (2010) Understanding the difficulty of training deep
  feedforward neural networks. In: AISTATS, vol~9, pp 249--256

\bibitem[{He et~al(2016)He, Zhang, Ren, and Sun}]{he2016deep}
He K, Zhang X, Ren S, Sun J (2016) Deep residual learning for image
  recognition. In: Proceedings of the IEEE Conference on Computer Vision and
  Pattern Recognition, pp 770--778

\bibitem[{Hornik et~al(1989)Hornik, Stinchcombe, and
  White}]{hornik1989multilayer}
Hornik K, Stinchcombe M, White H (1989) Multilayer feedforward networks are
  universal approximators. Neural networks 2(5):359--366

\bibitem[{Ioffe and Szegedy(2015)}]{ioffe2015batch}
Ioffe S, Szegedy C (2015) Batch normalization: Accelerating deep network
  training by reducing internal covariate shift. In: Proceedings of The 32nd
  International Conference on Machine Learning, pp 448--456

\bibitem[{Jia et~al(2014)Jia, Shelhamer, Donahue, Karayev, Long, Girshick,
  Guadarrama, and Darrell}]{jia2014caffe}
Jia Y, Shelhamer E, Donahue J, Karayev S, Long J, Girshick R, Guadarrama S,
  Darrell T (2014) Caffe: Convolutional architecture for fast feature
  embedding. arXiv preprint arXiv:14085093

\bibitem[{Kr{\"a}henb{\"u}hl et~al(2016)Kr{\"a}henb{\"u}hl, Doersch, Donahue,
  and Darrell}]{krahenbuhl2015data}
Kr{\"a}henb{\"u}hl P, Doersch C, Donahue J, Darrell T (2016) Data-dependent
  initializations of convolutional neural networks. In: International
  Conference on Learning Representations

\bibitem[{Krizhevsky and Hinton(2009)}]{krizhevsky2009learning}
Krizhevsky A, Hinton G (2009) Learning multiple layers of features from tiny
  images. Tech. rep., University of Toronto

\bibitem[{LeCun et~al(1989)LeCun, Boser, Denker, Henderson, Howard, Hubbard,
  and Jackel}]{lecun1989backpropagation}
LeCun Y, Boser B, Denker JS, Henderson D, Howard RE, Hubbard W, Jackel LD
  (1989) Backpropagation applied to handwritten zip code recognition. Neural
  computation 1(4):541--551

\bibitem[{LeCun et~al(1998{\natexlab{a}})LeCun, Bottou, Bengio, and
  Haffner}]{lecun1998gradient}
LeCun Y, Bottou L, Bengio Y, Haffner P (1998{\natexlab{a}}) Gradient-based
  learning applied to document recognition. Proceedings of the IEEE
  86(11):2278--2324

\bibitem[{LeCun et~al(1998{\natexlab{b}})LeCun, Bottou, Orr, and
  M{\"u}ller}]{lecun1998efficient}
LeCun Y, Bottou L, Orr GB, M{\"u}ller KR (1998{\natexlab{b}}) Efficient
  backprop. In: Neural Networks: Tricks of the Trade, pp 9--50

\bibitem[{Lin et~al(2014)Lin, Chen, and Yan}]{lin2014network}
Lin M, Chen Q, Yan S (2014) Network in network. In: International Conference on
  Learning Representations

\bibitem[{Mairal(2016)}]{mairal2016}
Mairal J (2016) End-to-end kernel learning with supervised convolutional kernel
  networks. In: Lee DD, Sugiyama M, Luxburg UV, Guyon I, Garnett R (eds)
  Advances in Neural Information Processing Systems 29, Curran Associates,
  Inc., pp 1399--1407

\bibitem[{Mairal et~al(2014)Mairal, Koniusz, Harchaoui, and
  Schmid}]{mairal2014convolutional}
Mairal J, Koniusz P, Harchaoui Z, Schmid C (2014) Convolutional kernel
  networks. In: Advances in Neural Information Processing Systems, pp
  2627--2635

\bibitem[{Mishkin and Matas(2016)}]{mishkin2015all}
Mishkin D, Matas J (2016) All you need is a good init. In: International
  Conference on Learning Representations

\bibitem[{Russakovsky et~al(2015)Russakovsky, Deng, Su, Krause, Satheesh, Ma,
  Huang, Karpathy, Khosla, Bernstein, Berg, and Fei-Fei}]{ILSVRC15}
Russakovsky O, Deng J, Su H, Krause J, Satheesh S, Ma S, Huang Z, Karpathy A,
  Khosla A, Bernstein M, Berg AC, Fei-Fei L (2015) {ImageNet Large Scale Visual
  Recognition Challenge}. International Journal of Computer Vision (IJCV)
  115(3):211--252

\bibitem[{Sch{\"o}lkopf and Smola(2002)}]{Scholkopf_2002}
Sch{\"o}lkopf B, Smola A (2002) Learning with Kernels- Support Vector Machines,
  Regularization, Optimization and Beyond. MIT Press, Cambridge, MA

\bibitem[{Serre et~al(2007)Serre, Wolf, Bileschi, Riesenhuber, and
  Poggio}]{serre2007robust}
Serre T, Wolf L, Bileschi S, Riesenhuber M, Poggio T (2007) Robust object
  recognition with cortex-like mechanisms. IEEE transactions on pattern
  analysis and machine intelligence 29(3):411--426

\bibitem[{Srivastava et~al(2014)Srivastava, Hinton, Krizhevsky, Sutskever, and
  Salakhutdinov}]{srivastava2014dropout}
Srivastava N, Hinton GE, Krizhevsky A, Sutskever I, Salakhutdinov R (2014)
  Dropout: a simple way to prevent neural networks from overfitting. Journal of
  Machine Learning Research 15(1):1929--1958

\bibitem[{Szegedy et~al(2015)Szegedy, Liu, Jia, Sermanet, Reed, Anguelov,
  Erhan, Vanhoucke, and Rabinovich}]{szegedy2015going}
Szegedy C, Liu W, Jia Y, Sermanet P, Reed S, Anguelov D, Erhan D, Vanhoucke V,
  Rabinovich A (2015) Going deeper with convolutions. In: Proceedings of the
  IEEE Conference on Computer Vision and Pattern Recognition, pp 1--9

\bibitem[{Williams and Hinton(1986)}]{williams1986learning}
Williams D, Hinton G (1986) Learning representations by back-propagating
  errors. Nature 323:533--536

\end{thebibliography}

\end{document}